\documentclass[11pt,a4paper]{article}
\usepackage[hyperref]{naaclhlt2019}
\usepackage{times}
\usepackage{latexsym}
\usepackage{graphicx}
\usepackage{tabularx, enumitem,booktabs, cellspace, multirow}
\usepackage{url,relsize}
\usepackage{subfig}
\usepackage{amsthm}
\usepackage{amsmath}
\usepackage{color}
\usepackage{amssymb}
\usepackage{subfiles}

\usepackage[hang]{footmisc}
\theoremstyle{definition}

\theoremstyle{definition}
\newtheorem{example}{Example}

\aclfinalcopy % Uncomment this line for the final submission
\def\aclpaperid{1354} %  Enter the acl Paper ID here

\newcommand{\openie}{{\textsc{\small OpenIE}}}
\newcommand{\neuron}{{\textsc{\small NeurON}}}
\newcommand{\neuralopenie}{{\textsc{\small NeuralOpenIE}}}
\newcommand{\bilstmcrf}{{\textsc{\small BiLSTM-CRF}}}

\newcommand{\amazonqa}{{\textit{AmazonQA}}}
\newcommand{\trustyou}{{\textit{ConciergeQA}}}

\newcommand{\cqa}{{\textsc{\small cQA}}}

\DeclareMathOperator*{\argmax}{argmax}

\newcommand{\nikita}[1]{#1}

\newcommand{\tagh}{$\langle \mbox{\em arg}_1\rangle$} % <arg1>
\newcommand{\tagH}{$\langle /\mbox{\em arg}_1\rangle$} % </arg1>
\newcommand{\tagr}{$\langle \mbox{\em rel}\rangle$} % <rel>
\newcommand{\tagR}{$\langle /\mbox{\em rel}\rangle$} % </rel>
\newcommand{\tagt}{$\langle \mbox{\em arg}_2\rangle$} % <arg2>
\newcommand{\tagT}{$\langle /\mbox{\em arg}_2\rangle$} % </arg2>

\title{Open Information Extraction from Question-Answer Pairs}

\author{Nikita Bhutani \thanks{\nikita{Part of the work was done while the author was at Megagon Labs.}}\\
  University of Michigan \\
  {\tt nbhutani@umich.edu} \\
\And
  Yoshihiko Suhara \\
  Megagon Labs \\
  {\tt yoshi@megagon.ai} \\\And
  Wang-Chiew Tan \\
  Megagon Labs \\
  {\tt wangchiew@megagon.ai} \\\AND
  Alon Halevy \\
  Megagon Labs \\
  {\tt alon@megagon.ai} \\\And
  H. V. Jagadish \\
  University of Michigan \\
  {\tt jag@eecs.umich.edu}
  }

\date{}

\begin{document}
\maketitle

\begin{abstract}
%Applications requiring deep understanding capabilities have led %to a renaissance of knowledge bases (KBs) in recent years. 

Open Information Extraction (\openie{}) extracts meaningful structured tuples from free-form text. Most previous work on \openie{} considers extracting data from one sentence at a time. We describe \neuron{}, a system for extracting tuples from question-answer pairs. Since real questions and answers often contain precisely the information that users care about, such information is particularly desirable to extend a knowledge base with. 
 
 \neuron{} addresses several challenges. First, an answer text is often hard to understand without knowing the question, and second, relevant information can span multiple sentences. 
To address these, \neuron{} formulates extraction as a multi-source sequence-to-sequence learning task, wherein it combines distributed representations of a question and an answer to generate knowledge facts. We describe experiments on two real-world datasets that demonstrate that \neuron{} can find a significant number of new and interesting facts to extend a knowledge base compared to state-of-the-art \openie{} methods.

\end{abstract}
\section{Introduction}

%wangchiew: move it to later part of the paper perhaps.
%Knowledge bases (KBs), such as DBpedia~\cite{auer2007dbpedia}, %Yago~\cite{hoffart2013yago2}, Wikidata~\cite{vrandevcic2014wikidata}, contain %large collections of facts about real-world entities and relations in a %machine-readable format. These KBs are useful for various tasks, including %question-answering~\cite{berant2013semantic} and semantic %parsing~\cite{berant2014semantic}. Although massive, most KBs are far from %complete. New knowledge is always emerging rapidly, making it difficult to %construct a comprehensive KB. The task of open information extraction %(\openie{}) — extracting structured relation tuples from plain text with no %specification of relations - is an attempt to alleviate the problem of such %knowledge sparsity.

Open Information Extraction (\openie) \cite{Banko:2007:IJCAI} is the problem 
of extracting structured data from a text corpus, without knowing a priori which relations will be extracted.  It is one of the primary technologies used in building knowledge bases (KBs) that, in turn, power question
answering~\cite{berant2013semantic}. The vast majority of previous work on 
\openie{} extracts structured information (e.g., triples) from 
 individual sentences. 
 
This paper addresses the problem of extracting structured data from conversational question-answer (\cqa) data. Often, \cqa{} data contains precisely the knowledge that users care about. As such, this data  offers a goal-directed method for extending existing knowledge bases. 
Consider, for example, a KB about a hotel that is used to power its website and/or a conversational interface for hotel guests. The KB  provides information about the hotel's services: complimentary breakfast, free wifi, spa. However, it may not include information about the menu/times for the breakfast, credentials for the wifi, or the cancellation policy for a spa appointment at the hotel. Given the wide range of information that may be of interest to guests, it is not clear how to extend the KB in the most effective way. However, the conversational logs, which many hotels keep, contain the actual questions from guests, and can therefore be used as a resource for extending the KB. Following examples illustrate the kind of data we aim to extract:

\begin{example}\label{ex:qapair1}
\noindent
\textit{Q: Does the hotel have a gym?}

\noindent
\textit{A: It is located on the third floor and is 24/7.}

\noindent
\textit{Tuple: $\langle$gym, is located on, third floor$\rangle$}
\end{example}

\begin{example}\label{ex:qapair2}
\noindent
\textit{Q: What time does the pool open?}

\noindent
\textit{A: 6:00am daily.}

\noindent
\textit{Tuple: $\langle$pool, open, 6:00am daily$\rangle$}

\end{example}
As can be seen from these examples, harvesting facts from \cqa{} data presents significant challenges. In particular, the system must interpret information collectively between the questions and answers. In this case, it must realize that
\textit{`third floor'} refers to the location of the  \textit{`gym'} and that 
 \textit{6:00am} refers to the opening time of the pool.
 \openie{} systems that operate over individual sentences ignore the discourse and context in a QA pair. Without knowing the question, they either fail to or incorrectly interpret the answer.

%%%%%%%%%%%%%%%%%%%%%%%%

This paper describes \neuron{},  an end-to-end system  for extracting information from \cqa{} data. We cast \openie{} from \cqa{} as a multi-source sequence-to-sequence generation problem to explicitly model both the question and answer in a QA pair. We propose a multi-encoder, constrained-decoder framework that uses two encoders to encode each of the question and answer to an internal representation. The two representations are then used by a decoder to generate an output sequence corresponding to an extracted tuple. For example, the output sequence of Example~\ref{ex:qapair2} is:

%\small
%\tagh{} pool \tagEND{} \tagr{} open \tagEND{} \tagt{} 6:00am daily \tagEND{} % 
% Yoshi: should we use the full close-tags (which makes the example in two lines)?
{
\noindent
%\small
\fontsize{8pt}{8pt}\selectfont
\tagh{} {\small pool} \tagH{}\tagr{} open \tagR{}\tagt{} 6:00am daily \tagT{} 
}

While encoder-decoder frameworks have been used extensively for machine translation and summarization, there are two key technical challenges in extending them for information extraction from \cqa{} data. First, it is vital for the translation model to learn constraints such as, arguments and relations are sub-spans from the input sequence, output sequence must have a valid syntax (e.g., \tagh{} must precede \tagr{}). These and other constraints can be integrated as \textit{hard} constraints in the decoder. Second, the model must recognize auxiliary information that is irrelevant to the KB. For example, in the hotel application, \neuron{} must learn to discard greetings in the data. Since existing facts in the KB are representative of the domain of the KB, this prior knowledge can be incorporated as \textit{soft} constraints in the decoder to rank various output sequences based on their relevance. 
Our contributions are summarized below:

\setlist[itemize]{leftmargin=*}
\begin{itemize}[noitemsep,topsep=0pt]
\item We develop \neuron{}, a system for extracting information from \cqa{} data. \neuron{} is a novel multi-encoder constrained-decoder method that explicitly models both the question and the answer of a QA pair. It incorporates vocabulary and syntax as \textit{hard} constraints and prior knowledge as \textit{soft} constraints in the decoder. 
% \item We conduct comprehensive experiments on two real-world \cqa{} datasets. Experimental results demonstrate that our proposed approach can find additional 15\%-22\% tuples, not extracted by state-of-the-art sentence-based models. 
\item We conduct comprehensive experiments on two real-world \cqa{} datasets. Our experimental results show that the use of hard and soft constraints improves the extraction accuracy and \neuron{} achieves the highest accuracy in extracting tuples from QA pairs compared with state-of-the-art sentence-based models, with a relative improvement as high as 13.3\%.
%by as much as 5.4\%. 
\neuron's higher accuracy and ability to discover 15-25\% tuples that are not extracted by state-of-the-art models make it suitable as a tuple extraction tool for KB extension.
\item We present a case study to demonstrate how a KB can be extended iteratively using tuples extracted using \neuron{}. In each iteration, only relevant tuples are included in the KB. In turn, the extended KB is used to improve relevance scoring for subsequent iterations.
% We conduct comprehensive experiments to compare the proposed method to baseline algorithms to demonstrate their effectiveness in extending KBs. \wctan{be more concrete once the experiments are done}
\end{itemize}

\begin{figure*}[ht]
  \center
  \includegraphics[width=0.85\linewidth]{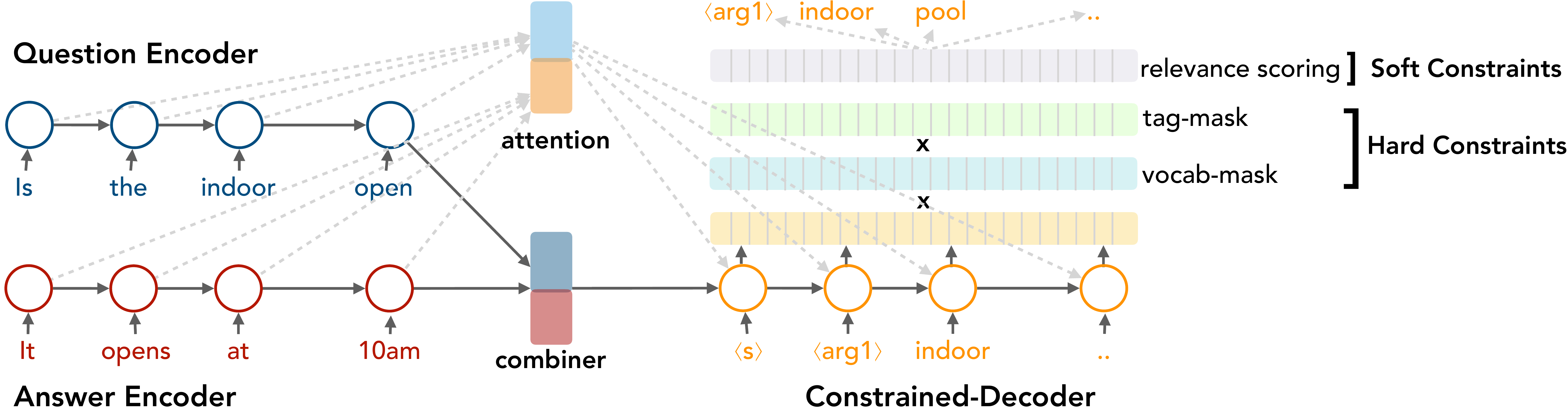}
  \caption{Multi-Encoder, Constrained-Decoder model for tuple extraction from $(q,a)$. 
}
  \label{fig:architecture}
\end{figure*}

\iffalse
\noindent
{\bf Outline:} We introduce key terminology and task formulation in Sec.~\ref{sec:definition}. We present our methodology for extraction in Sec.~\ref{sec:technical} and experimental results in Sec.~\ref{sec:experiments}, followed by 
a case study on constructing KB in Sec.~\ref{sec:extension}. We describe related work in Sec.~\ref{sec:relatedwork} and conclude in Sec.~\ref{sec:conclusion}.
\fi

\section{Task Formulation}
\label{sec:definition}

In this work, we choose to model an \openie{} extraction from a question-answer (QA) pair as a tuple consisting of a single relation with two arguments, where the relation and arguments are contiguous spans from the QA pair. Formally, let $(q,a)$ be a QA pair, where question $q = (q_1, q_2, ..., q_m)$ and answer $a = (a_1, a_2, ..., a_n)$ are word sequences.  The output is a triple ($arg_1$,$rel$,$arg_2$) extracted
from $(q,a)$. The output triple can be naturally interpreted as a sequence $y = (y_1, y_2, ..., y_o)$
%Formally, the triple is also a sequence $y = (y_1, y_2, ..., y_o)$ 
where $y_i$ is either a word or a placeholder tag (\tagh, \tagr, \tagt) that marks relevant portions of the triple. 
%Thus, the vocabulary of $y$ is restricted to vocabulary of $(q,a)$ and placeholder tags.
In \openie{}, the extracted tuple should be asserted by the input QA pair. Formulating this, therefore, requires the vocabulary of $y$ to be restricted to the vocabulary of $(q,a)$ and placeholder tags.

Following this definition, our aim is to directly model the conditional probability $p(y|q,a)$ of mapping input sequences $q$ and $a$ into an output sequence:

%\vspace{-5pt}
{
\setlength{\abovedisplayskip}{0pt}%
\setlength{\belowdisplayskip}{0pt}%
\setlength{\abovedisplayshortskip}{-2pt}%
\setlength{\belowdisplayshortskip}{-2pt}%
\begin{equation}\label{eq:output_seq_probability}
    P(y|q,a) = \prod_{i=1}^{o} p(y_i | y_1,\dots, y_{i-1}, q, a).
\end{equation}
}
%\vspace{-5pt}

\nikita{In our formulation, a triple is generated as a sequence: a \textit{head} argument phrase $arg_1$, followed by a relation phrase \textit{rel} and a \textit{tail} argument phrase $arg_2$. It is possible to consider different orderings in the output sequence (such as ($rel$,$arg_1$,$arg_2$)). However, the goal of \openie{} is to identify the relation phrase that holds between a pair of arguments. Our representation is, thus, consistent with this definition as it models the relation phrase to depend on the head argument.}

%Both the input and the output sequences can have different lengths. 
%for argument and relation boundaries (e.g., %\tagh{}, \tagr{}). 
%Neither the arguments nor the relation in the output sequence are specified in advance. 

%\section{Tuple extraction}\label{sec:technical}
\section{NeurON: A Multi-Encoder Constrained-Decoder Model}\label{sec:technical}

{\bf Overview of \neuron~} We propose to extract tuples using a variation of an encoder-decoder RNN architecture~\cite{cho2014learning} operating on variable-length sequences of tokens. Fig.~\ref{fig:architecture} shows the architecture of \neuron{}. It uses two encoders to encode question and answer sequences in a QA pair separately into fixed-length vector representations. A decoder then decodes the vector representations into a variable-length sequence corresponding to the tuple. The decoder is integrated with a set of \textit{hard constraints} (e.g., output vocabulary) and \textit{soft constraints} (e.g., relevance scoring) suited for the extraction task.
%\subsection{Multi-Encoder Constrained-Decoder Model}

\subsection{Multiple Encoders}

Given an input QA pair, two RNN encoders separately encode the question and answer. The question encoder converts $q$ into hidden representation $h^q=(h^q_1,...,h^q_m)$ and the answer encoder converts $a$ into $h^a=(h^a_1,...,h^q_n)$, where $h^q_t= \text{lstm}(q_t, h^q_{t-1})$ is a non-linear function represented by the long short-term memory (LSTM) cell. The combiner combines the encoders' states and initializes the hidden states $\mathbf{h}$ for the decoder: 

%\vspace{-5pt}
{
\setlength{\abovedisplayskip}{0pt}%
\setlength{\belowdisplayskip}{0pt}%
\setlength{\abovedisplayshortskip}{-2pt}%
\setlength{\belowdisplayshortskip}{-2pt}%
\begin{equation*}
% 	\mathbf{h} = \mathrm{tanh}(W_c [\mathbf{h^q} \circ
%\mathbf{h^a}] ).
h = \mathrm{tanh}(W_c [{h^q} \circ {h^a}] ),
\end{equation*}
}
%\vspace{-5pt}

\noindent where $\circ$ denotes concatenation. The decoder stage uses the hidden states to generate the output $y$ with another LSTM-based RNN. The probability of each token is defined as:

%\vspace{-5pt}
{
\setlength{\abovedisplayskip}{0pt}%
\setlength{\belowdisplayskip}{0pt}%
\setlength{\abovedisplayshortskip}{-2pt}%
\setlength{\belowdisplayshortskip}{-2pt}%
\begin{equation}\label{eq:output_probability}
	p(y_t) = \text{softmax}((s_t \circ c^q_t \circ c^a_t) W_y),
\end{equation}
}
%\vspace{-5pt}

\noindent
where $s_t$ denotes the decoder state, $s_0=h$ and $s_t=\text{lstm}((y_{t-1} \circ c^q_t \circ c^a_t)W_s, s_{t-1})$. The decoder is initialized by the last hidden state from the combiner. It uses the previous output token at each step. Both $W_y$ and $W_s$ are learned matrices. Each decoder state is concatenated with \textit{context vectors} derived from the hidden states of the encoders. Context vector $c_t$ is the weighted sum of the encoder hidden states, i.e. $c^q_t = \sum_{i=1}^m \alpha_{t_i}h^q_i$, where $\alpha_{t_i}$ corresponds to an attention weight. The attention model~\cite{bahdanau2014neural} helps the model learn to focus on specific parts of the input sequences, instead of solely relying on hidden vectors of the decoders' LSTM. This is crucial for extraction from $(q,a)$ pairs where input sequences tend to be long.

\subsection{Constrained Decoder}

The decoder finds the best hypothesis (i.e., the best output sequence) for the given input representations. Typically, the output sequence is generated, one unit at a time, using \textit{beam search}. At each time step, the decoder stores the top-$k$ scoring partial sequences, considers all possible single token extensions of them, and keeps $k$ most-likely sequences based on model's probabilities (Eq.~\ref{eq:output_seq_probability}). As soon as the $\langle /S \rangle$ symbol is appended, the sequence is removed from the beam and added to the set of complete sequences. The most-likely complete sequence is finally generated.

\smallskip
\noindent
{\bf Hard Constraints~} \nikita{While such encoder-decoder models typically outperform conventional approaches \cite{cho2014learning,zoph2016multi,xiong2016dynamic} on a wide variety of tasks including machine translation and question answering, the accuracy and training efficiency has been shown to improve when the model is integrated with the constraints of the output domain \cite{xiao2016sequence,yin2017syntactic}. Motivated by these, \neuron{} 
allows constraints relevant to information extraction to be incorporated in the model.} % applies constraints relevant to information extraction. Specifically, it enforces vocabulary and structural constraints on the output.
Specifically, we describe how the decoder can enforce vocabulary and structural constraints on the output.

\begin{figure}
  \center
\includegraphics[width=1\linewidth]{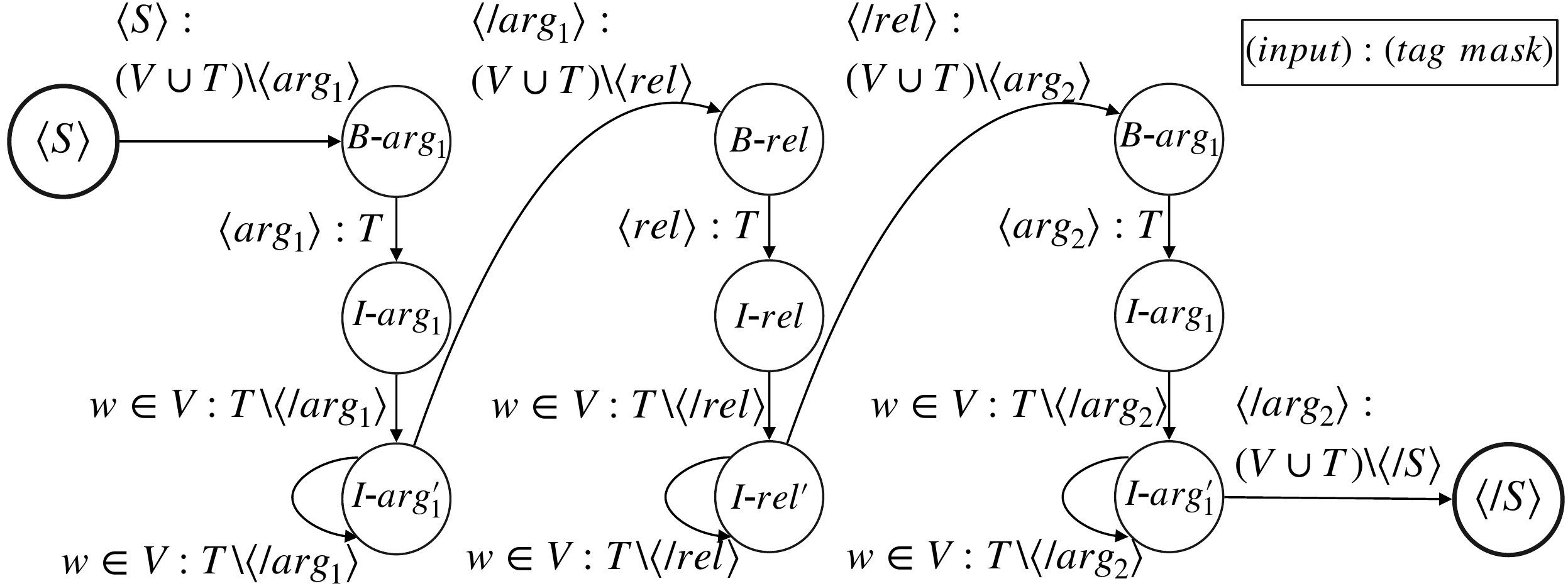}
%\caption{Rules for masking outputs based on placeholder tag states. $V$ is is the vocabulary including placeholder tags, $T$ is the set of placeholder tags.
% The left and right elements of $:$ denote input and tag mask $r$ of each transition respectively.
\caption{State diagram for tag masking rules. $V$ is the vocabulary including placeholder tags, $T$ is the set of placeholder tags.
% \nikita{Replace with high-quality image if this is easier to understand than the table.}
}
  \label{fig:rules}
  %\vspace{-6pt}
\end{figure}

\begin{itemize}[leftmargin=*,wide = 0pt]
    \item {\bf Vocabulary constraints.~} Since the arguments and relations in the extracted tuples typically correspond to the input QA pair, the decoder must constraint the space of next valid tokens when generating the output sequence. \neuron{} uses a masking technique in the decoder to mask the probability of tokens (as in Eq.~\ref{eq:output_probability}) that do not appear in the input $(q,a)$ pair. Specifically, it computes a binary mask vector $v$, where $|v|$ is vocabulary size and $v_i = 1$ if and only if $i$-th token appears in $q$ or $a$. 
    %iff $i \in (q,a)$. 
    %
    The probability of each token is modified as:

\vspace{-5pt}
{
\setlength{\abovedisplayskip}{0pt}%
\setlength{\belowdisplayskip}{0pt}%
\setlength{\abovedisplayshortskip}{-2pt}%
\setlength{\belowdisplayshortskip}{-2pt}%
\begin{equation}
 	p(y_t) = \text{softmax}((s_t \circ c^q_t \circ c^a_t) W_y \otimes v) ,
 \end{equation}
 }
 \vspace{-5pt}
 
 \noindent
where $\otimes$ indicates element-wise multiplication.

\item
{\bf Structural constraints.~} For the output sequence to correspond to a valid tuple with non-empty arguments, the decoding process must conform to the underlying grammar of a tuple. For instance, decoding should always begin in the $\langle S \rangle$ state, where only $\langle arg_1 \rangle$ can be generated. In subsequent time steps, all other placeholders except $\langle /arg_1 \rangle$ should be restricted to ensure a non-empty argument. Once $\langle /arg_1 \rangle$ is generated, $\langle rel \rangle$ must be generated in the next time step and so on. The various states and grammar rules can be described as a finite state transducer as shown in Figure~\ref{fig:rules}.

%Table~\ref{tab:rules}. 
%
Depending upon the state, \neuron{} generates a mask $r$ based on this grammar and uses $r$ to further modify the probabilities of the tokens as follows:

\vspace{-5pt}
{
\setlength{\abovedisplayskip}{0pt}%
\setlength{\belowdisplayskip}{0pt}%
\setlength{\abovedisplayshortskip}{0pt}%
\setlength{\belowdisplayshortskip}{0pt}%
\begin{equation}
 	p(y_t) = \text{softmax}((s_t \circ c^q_t \circ c^a_t) W_y \otimes v \otimes r ) .
 \end{equation}
 }
 \vspace{-5pt}
 
\end{itemize}

% \setlength{\tabcolsep}{4pt} % Default value: 6pt
% \begin{table}
% \small
% 	\begin{tabularx}{0.50\textwidth}{|l|X|X|X|}
%     	\toprule
%         current state	&	token to mask	&	next token & next state\\
%         \toprule
%   			$begin\_arg_1$	&	$ V \backslash \langle arg_1 \rangle $	&	$\langle arg_1 \rangle$ & $in\_arg_1$\\
%   		\midrule 
%   		    \multirow{2}{*}{$in\_arg_1$}&\multirow{2}{*}{$T\backslash\langle /arg_1 \rangle$}&
%   		 	$V \backslash \langle /arg_1 \rangle $ & $in\_arg_1$\\
%   		 	\cline{3-4}
%   		    & &  $\langle /arg_1 \rangle$ & $begin\_rel$\\
%   		\toprule
%   			$begin\_rel$	&	$ V \backslash \langle rel \rangle $	&	$\langle rel \rangle$ & $in\_rel$\\
%   		\midrule 
%   		\multirow{2}{*}{$in\_rel$}&\multirow{2}{*}{$T \backslash\langle /rel \rangle$}&
%   		 	$V \backslash \langle /rel \rangle $ & $in\_rel$\\
%   		 	\cline{3-4}
%   		    & &  $\langle /rel \rangle$ & $begin\_arg_2$\\
%   		\toprule
%   		    $begin\_arg_2$	&	$ V \backslash \langle arg_2 \rangle $	&	$\langle arg_2 \rangle$ & $in\_arg_2$\\
%   		\midrule
%   			\multirow{2}{*}{$in\_arg_2$}&\multirow{2}{*}{$T \backslash\langle /arg_2 \rangle$}&
%   		 	$V \backslash \langle /arg_2 \rangle $ & $in\_arg_2$\\
%   		 	\cline{3-4}
%   		    & &  $\langle /arg_2 \rangle$ & $end\_arg_2$\\
%   		\bottomrule
% 	\end{tabularx}
%   	\caption{Rules for masking outputs based on placeholder tag states. $V$ is is the vocabulary including placeholder tags, $T$ is the set of placeholder tags.}
%   	\label{tab:rules}
% \end{table}

%\subsection{Tuple Relevance Scoring}
\smallskip
\noindent
{\bf Soft Constraints~} 
\nikita{\openie{} systems are typically used to extract broad-coverage facts to extend existing KBs. Facts already existing in the KB are representative of the domain of the KB. It is, therefore, useful to incorporate this prior knowledge in the extraction itself. \neuron{} is able to use prior knowledge (incorporated as soft constraints) in the decoder to understand the relevance of candidate extractions and adjust the ranking of various output sequences accordingly. To see why such soft constraints can be useful, consider the example:}

\begin{example}\label{ex:qapair4}
\noindent
\textit{Q: ``Is the pool open?''}

\noindent
\textit{A: ``I am sorry but our pool reopens at 7:00am.''}

\noindent
\textit{Tuple: $\langle$I, am, sorry$\rangle$; $\langle$pool, reopens at, 7:00am$\rangle$}
\end{example}

\noindent
Both the tuple facts are correct given the input QA pair but only the second tuple contains useful information. Filtering such irrelevant facts is difficult without additional evidence. 

The multi-encoder and constrained-decoder in \neuron{} are jointly optimized to maximize the log probability of output sequence conditioned on the input sequences. At inference, the decoder estimates the likelihood of various candidate sequences and generates the sequence with the highest likelihood. As shown in Eq.~\ref{eq:output_seq_probability}, this likelihood is conditioned solely on the input $(q,a)$ pair, thus
increasing the possibility of obtaining facts that may be correct but irrelevant. Instead, if a relevance scoring function were integrated at extraction time, the candidate output sequences could be re-ranked so that the predicted output sequence is likely to be both correct and relevant.

Learning a relevance scoring function can be modeled as a KB completion task, where missing facts have to be inferred from existing ones. A promising approach is to learn vector representations of entities and relations in a KB by maximizing the total plausibility of existing facts in the KB~\cite{wang2017knowledge}. For a new candidate output sequence, its plausibility can be predicted using the learned embeddings for the entities and relation in the sequence.

In \neuron{}, we learn the entity and relation embeddings using Knowledge Embedding (KE) methods such as TransE~\cite{bordes2013translating} and HolE~\cite{nickel2016holographic}. Note that \neuron{} is flexible with how the relevance scoring function is learned or which KE method is chosen. In this paper, we use TransE for evaluation. TransE computes the plausibility score $S$ of a tuple $y=\langle arg_1, rel, arg_2 \rangle$ as:

%\vspace{-5pt}
{
\setlength{\abovedisplayskip}{0pt}%
\setlength{\belowdisplayskip}{0pt}%
\setlength{\abovedisplayshortskip}{0pt}%
\setlength{\belowdisplayshortskip}{0pt}%
\begin{equation*}
	S(y)  = || \mathbf{v}_{arg_1} + \mathbf{v}_{rel} - \mathbf{v}_{arg_2}||,
\end{equation*}
}
%\vspace{-5pt}

\noindent
where $\mathbf{v}_{arg_1}$, $\mathbf{v}_{rel}$, and $\mathbf{v}_{arg_2}$ are embedding vectors for $arg_1$, $rel$, and $arg_2$ respectively. Following~\cite{jain2018mitigating}, we compute the embedding vectors of out-of-vocabulary arguments (and relations) as the average of embedding vectors of known arguments (and relations). We generate the most-likely output based on its conditional probability and plausibility score: 

%\vspace{-10pt}
{
\setlength{\abovedisplayskip}{3pt}%
\setlength{\belowdisplayskip}{6pt}%
\setlength{\abovedisplayshortskip}{0pt}%
\setlength{\belowdisplayshortskip}{0pt}%
\begin{equation}\label{eq:objective_func}
	\hat{y} = \argmax_y (\log P(y| q, a) + \gamma \log S(y)).
\end{equation}
}

To implement the tuple relevance scoring function, we employ the re-ranking approach, which is a common technique for sequence generation methods~\cite{luong2015effective}. Our re-ranking method first obtains candidates from a beam search decoder and then re-ranks the candidates based on the objective function (Eq.~\ref{eq:objective_func}).

\section{Experiments}\label{sec:experiments}
% \wctan{provide overview of results}
We evaluated the performance of \neuron{} on two \cqa{} datasets. In our analysis, we find that integrating hard and soft constraints in the decoder improved the extraction performance irrespective of the number of encoders used. Also, 15-25\% of the tuples extracted by \neuron{} were not extracted by state-of-the-art sentence-based methods. 

\subsection{Datasets and Data Preparation}

{\bf \trustyou{}} is a real-world internal corpus of 33,158 QA pairs collected via a multi-channel communication platform for guests and hotel staff. Questions (answers) are always made by guests (staff). An utterance has 36 tokens on average, and there are 25k unique tokens in the dataset. \nikita{A QA utterance has 2.8 sentences on average, with the question utterance having 1.02 sentences on average and answer utterance having 1.78 sentences on average.}

\noindent
{\bf \amazonqa{}}~\cite{wan2016modeling,mcauley2016addressing} is a public dataset with 314,264 QA pairs about electronic products on \textit{amazon.com}. The dataset contains longer and more diverse utterances than the \trustyou{} dataset: utterances have an average of 45 tokens and the vocabulary has more than 50k unique tokens. \nikita{A QA utterance has 3.5 sentences on average. The question utterances had 1.5 sentences on average and the answer having 2 sentences.}

For training \neuron{}, we bootstrapped a large number of high-quality training examples using a state-of-the-art \openie{} system. Such bootstrapping has been shown to be effective in information extraction tasks~\cite{schmitz2012open,saha2017bootstrapping}. 
The StanfordIE~\cite{angeli2015leveraging} system is used to
extract tuples from QA pairs for training examples. To further obtain high-quality tuples, we filtered out tuples that occur too infrequently ($<5$) or too frequently ($>100$). For each tuple in the set, we retrieved all QA pairs that contain all the content words of the tuple and included them in the training set. This helps create a training set encapsulating the multiplicity of ways in which tuples are expressed across QA pairs. \nikita{We randomly sampled 100 QA pairs from our bootstrapping set and found 74 of them supported the corresponding tuples. We find this quality of bootstrapped dataset satisfactory, since the seed tuples for bootstrapping could be noisy as they were generated by another \openie{} system.}

Our bootstrapped dataset included training instances where a tuple matched (a) tokens in the questions exclusively, (b) tokens in the answers exclusively, (c) tokens from both questions and answers. Table~\ref{tab:train_data} shows the distribution of the various types of training instances. Less than 40\% (30\%) of ground truth tuples for \trustyou{} (\amazonqa{}) exclusively appear in the questions or answers. Also, 22.1\% (37.2\%) of ground truth tuples for \trustyou{} (\amazonqa{}) are extracted from the combination of questions and answers. These numbers support our motivation of extracting tuples from QA pairs.  We used standard techniques to construct training/dev/test splits so that QA pairs in the three sets are disjoint.  Table~\ref{tab:data} shows the details of the various subsets. 
%\yoshi{did we make test data and training data mutually exclusive wrt QA pairs (maybe not tuples.) How many ground truth tuples in the test dataset do not appear in the training data?} \nikita{The training data and test data have mutually exclusively QA pairs.} 

\begin{table}
    \fontsize{9}{11}\selectfont
	\begin{tabularx}{0.487\textwidth}{|X|l|l|}
    	\toprule
        Instance type	&	{\small \trustyou{}}	&	{\small \amazonqa{}} \\
        \toprule
  			Exclusively from question & 13.9$\%$ & 13.8$\%$ \\
  		\midrule 
  			Exclusively from answer & 25.8$\%$ & 17.6$\%$ \\
  		\midrule 
  			Ambiguous & 36.9$\%$ & 29.8$\%$ \\
  	    \midrule 
  			Jointly from Q-A & 23.4$\%$ & 38.8$\%$ \\
  		\bottomrule
	\end{tabularx}
  	\caption{Various types of training instances.}
  	\label{tab:train_data}
\end{table}

\setlength{\tabcolsep}{4pt}
\begin{table}
    %\small
    \fontsize{9}{11}\selectfont
	\begin{tabularx}{0.5\textwidth}{|X|l|l|l|l|l|}
    	\toprule
        Dataset	&	Q-A	&	$|$V$|$ &   Train	&	Dev	&	Test (Q-A)\\
        \toprule
  			\trustyou{}	&	33k	&   25k   &	1.25M	&	128k	&	2,905 \\
  		\midrule 
  			\amazonqa{}  & 314k  & 50k &   1.43M	&	159k	&	39,663 \\
  		\bottomrule
	\end{tabularx}
  	\caption{Training, Dev and Test splits.}
  	\label{tab:data}
  %\vspace{-5pt}
\end{table}

\subsection{Baseline Approaches}
We compared \neuron{} with two methods that can be trained for tuple extraction from QA pairs: \bilstmcrf{}~\cite{huang2015bidirectional} and \neuralopenie{}~\cite{cui2018neural}. \bilstmcrf{} is a sequence tagging model that has achieved state-of-the-art accuracy on POS, chunking, NER and \openie{} \cite{stanovsky2018supervised} tasks. For \openie{}, the model predicts boundary labels (e.g., B-ARG1, I-ARG1, B-ARG2, O) for the various tokens in a QA pair. \neuralopenie{} is an encoder-decoder model that generates a tuple sequence given an input sequence. Since it uses a single encoder, we generate the input sequence by concatenating the question and answer in a QA pair. We trained all the models using the same training data.

\subsection{Performance Metrics}

We examine the performance of different methods using three metrics: \textit{precision}, \textit{recall}, and \textit{relative coverage} (RC). Given a QA pair, each system returns a sequence. We label the sequence correct if it matches one of the ground-truth tuples for the QA pair, incorrect otherwise. We then measure precision of a method (i.e., \# of correct predictions of the method / \# of question-answer pairs) and recall (i.e., \# of correct predictions of the method / \# of correct predictions of any method) following \cite{Stanovsky2016EMNLP}. To compare the coverage of sequences extracted by \neuron{} against the baseline method, we compute relative coverage of \neuron{} as the fraction of all correct predictions that were generated exclusively by \neuron{}. Specifically,

{
\setlength{\abovedisplayskip}{0pt}%
\setlength{\belowdisplayskip}{0pt}%
\setlength{\abovedisplayshortskip}{0pt}%
\setlength{\belowdisplayshortskip}{0pt}%
\small
\begin{equation*}
	RC = \frac{|\text{TP}_{\small\neuron{}} \backslash \text{TP}_{baseline}|}{|\text{TP}_{\small\neuron{}} \bigcup \text{TP}_{baseline}|},
\end{equation*}
}
\normalsize

\noindent
where ${\rm TP}$ denotes the correct predictions.

\subsection{Model Training and Optimization}

We implemented \neuron{} using OpenNMT-tf~\cite{klein2017opennmt}
%\footnote{\url{https://github.com/OpenNMT/OpenNMT-tf}}
, an open-source neural machine translation system that supports multi-source encoder-decoder models. We implemented \neuralopenie{} using the same system. We used the open-source implementation of \bilstmcrf{}~\cite{Reimers:2017:EMNLP}.
%\footnote{\url{https://github.com/UKPLab/emnlp2017-bilstm-cnn-crf}}.
For fair comparison, we used identical configurations for \neuron{} and \neuralopenie{}. Each encoder used a 3-layer bidirectional LSTM and the decoder used a 3-layer bidirectional LSTM. The models used 256-dimensional hidden states, 300-dimensional word embeddings, and a vocabulary size of 50k. The word embeddings were initialized with pre-trained GloVe embeddings (glove.6B)~\cite{pennington2014glove}.
%\footnote{\url{http://nlp.stanford.edu/data/glove.6B.zip}}.
We used an initial learning rate of 1 and optimized the model with stochastic gradient descent. We used a decay rate of 0.7, a dropout rate of 0.3 and a batch size of 64. The models were trained for 1M steps for the \trustyou{} dataset and 100k steps for the \amazonqa{} dataset. We used TESLA K80 16GB GPU for training the models. We trained the KE models for relevance scoring using our bootstrapped training dataset. For integrating the relevance scoring function, we experimented with different values for $\gamma$ and found it not have a major impact within a range of 0.02 to 0.2. We used a value of 0.05 in all the experiments.

%In our experiments, we found $\gamma$ value of 0.05 worked well for integrating the relevance scoring function.

% \yoshi{Remove this paragraph?} The relevance scoring model is updated in each iteration based on the facts in the KB. In the first iteration, when there are no facts in the KB, we train the model based on the boostrapped dataset used to train the encoder-decoder model.

\begin{table}
%\fontsize{9}{11}\selectfont
\centering
	\begin{tabularx}{0.50\textwidth}{|X|l|l|l|}
    	\toprule
        	Method	&	P	& R & RC\\
        \toprule
  			\neuralopenie{} (baseline)	&	0.769	& 0.580 & - \\
            \hline
            \hspace{0.2em} + hard constraints	&	0.776	& 0.585  & - \\
            \hline
            \hspace{0.2em} + hard and soft constraints	&	0.796	& 0.600 & - \\
  		\midrule
        \midrule
  			\neuron{} (our method)	&	0.791	& 0.597 &	0.224\\
            \hline
            \hspace{0.2em} + hard constraints	&	0.792	& 0.597 & 0.204\\
            \hline
            \hspace{0.2em} + hard and soft constraints	& {\bf 0.807} & {\bf 0.608} & {\bf 0.245} \\
  		\bottomrule
	\end{tabularx}
  	\caption{Precision (P), Recall (R), and Relative Coverage (RC) results on \trustyou{}. }
  	\label{tab:trustyou_results}
\end{table}

\subsection{Experimental Results}

The \bilstmcrf{} model showed extremely low ($2\mbox{-}15\%$) precision values. Very few of the tagged sequences ($32\mbox{-}39\%$) could be converted to a tuple. Most tagged sequences had multiple relations and arguments, indicating that it is difficult to learn how to tag a sequence corresponding to a tuple. The model only learns how to best predict tags for each token in the sequence, and does not take into account the long-range dependencies to previously predicted tags. This is still an open problem and is outside the scope of this paper.

%While this issue can be alleviated by grouping tuples for each QA pair by relation head-word~\cite{stanovsky2018supervised}, extraction will additionally require including the index of the relation’s head in the QA pair.

Tables~\ref{tab:trustyou_results} and~\ref{tab:amazonqa_results} show the performance of \neuralopenie{} and \neuron{} on the two \cqa{} datasets. \neuron{} achieves higher precision on both the datasets. This is because \neuralopenie{} uses a single encoder to interpret the question and answer in the same vector space, which leads to lower performance. Furthermore, concatenating the question and answer makes the input sequence too long for the decoder to capture long-distance dependencies in history~\cite{zhang2016recurrent,toral2017multifaceted}. Despite the attention mechanism, the model ignores past alignment information. This makes it less effective than the dual-encoder model used in \neuron{}.

The tables also show that incorporating task-specific hard constraints helps further improve the overall precision and recall, regardless of the methods and the datasets. Re-ranking the tuples based on the soft constraints derived from the existing KB further improves the performance of both methods in \trustyou{} and \neuralopenie{} in \amazonqa{}. The existing KB also helps boost the likelihood of a correct candidate tuple sequence that was otherwise scored to be less likely. Lastly, we found that \neuron{} has significant relative coverage; it discovered significant additional, unique tuples missed by \neuralopenie{}. 

Table~\ref{tab:amazonqa_results} shows a slight
decrease in performance for \neuron{}
after soft constraints are added.
This is likely caused by the lower quality KE model due
to the larger vocabulary in \amazonqa. In contrast, 
even with the lower quality KE model, \neuralopenie{} improved slightly. This is likely because
the \neuralopenie{} model, at this stage, still had a larger margin for improvement.
We note however that learning the best KE model is not the focus of this work.

%into \neuron{} with hard constraints decreased the performance. 
%We consider the degradation was caused by the insufficient %quality of the KE model due to the large vocabulary size in %\amazonqa{}. As vocabulary size increases, the combination of %entities and relations increases in a quadratic manner. In %contrast, the same KE model improved the performance of %\neuralopenie{} since the \neuralopenie{} model had a large room %for improvement so even the {\it not-very-good} KE model could %improve the performance. Therefore, with a good quality KE %model, incorporating soft constraints improves the quality of %extraction regardless of the method. Note that learning the best %KE model is not the primary scope of this paper and one of our %future work.

\amazonqa{} is a more challenging dataset than \trustyou{}: longer utterances (avg. 45 tokens vs. 36 tokens) and richer vocabulary ($>50\mathrm{k}$ unique tokens vs. $<25\mathrm{k}$ unique tokens). This is reflected in lower precision and recall values of both the systems on the \amazonqa{} dataset. While the performance of end-to-end extraction systems depends on the complexity and diversity of the dataset, incorporating hard and soft constraints alleviates the problem to some extent.

% Tables~\ref{tab:trustyou_results} and~\ref{tab:amazonqa_results}
% also show that the performance of 
% end-to-end extraction systems is 
% dependent on the complexity and the diversity of the dataset. \amazonqa{} is a more challenging dataset than \trustyou{}: longer utterances (avg. 45 tokens vs. 36 tokens) and richer vocabulary ($>50\mathrm{k}$ unique tokens vs. $<25\mathrm{k}$ unique tokens).
% %Consequently, learning a high-quality model for the \amazonqa{} %dataset is significantly more difficult. 
% Incorporating hard constraints alleviates the problem to some extent, as is reflected in higher gains in precision and recall in \amazonqa{} dataset comparison to \trustyou{} dataset. However, since learning a good relevance scoring function over a complex dataset is difficult, incorporating soft constraints does not further boost performance.

\begin{table}
%\fontsize{9}{11}\selectfont
\centering
	\begin{tabularx}{0.50\textwidth}{|X|l|l|l|}
    	\toprule
        	Method	&	P		& R &	RC\\
        \toprule
  			\neuralopenie{} (baseline) &	0.557	& 0.594 & - \\
            \hline
            \hspace{0.2em} + hard constraints	&	0.563	& 0.601  & - \\
            \hline
            \hspace{0.2em} + hard and soft constraints	&	0.571 & 0.610  & -\\
  		\midrule
        \midrule
  			\neuron{} (our method) &	0.610	& 0.652 &	0.139\\
            \hline
            \hspace{0.2em} + hard constraints	& {\bf 0.631}	& {\bf 0.674} &	{\bf 0.164} \\
            \hline
            \hspace{0.2em} + hard and soft constraints	&	0.624	& 0.666 &	0.149\\
  		\bottomrule
	\end{tabularx}
  	\caption{Precision (P), Recall (R), and Relative Coverage (RC) results on \amazonqa{} dataset. }
  	\label{tab:amazonqa_results}
\end{table}

End-to-end extraction systems tend to outperform rule-based systems on extraction from \cqa{} datasets. We observed that training data for \trustyou{} had a large number ($>$ 750k) dependency-based pattern rules, of which $<$ 5\% matched more than 5 QA pairs. The set of rules is too large, diverse and sparse to train an accurate rule-based extractor. Even though our training data was generated by bootstrapping from a rule-based extractor StanfordIE, we found only 51.5\% (30.7\%) of correct tuples from \neuron{} exactly matched the tuples from StanfordIE in \trustyou{} (\amazonqa{}). This indicates that \neuron{} combined information from question and answer, otherwise not accessible to sentence-wise extractors. As an evidence, we found 11.4\% (6.1\%) of tuples were extracted from answers, 16.8\% (5.0\%) from questions, while 79.6\% (82.5\%) combined information from questions and answers in \trustyou{} (\amazonqa{}).

\begin{figure}[t]
\centering
\includegraphics[width=0.9\linewidth]{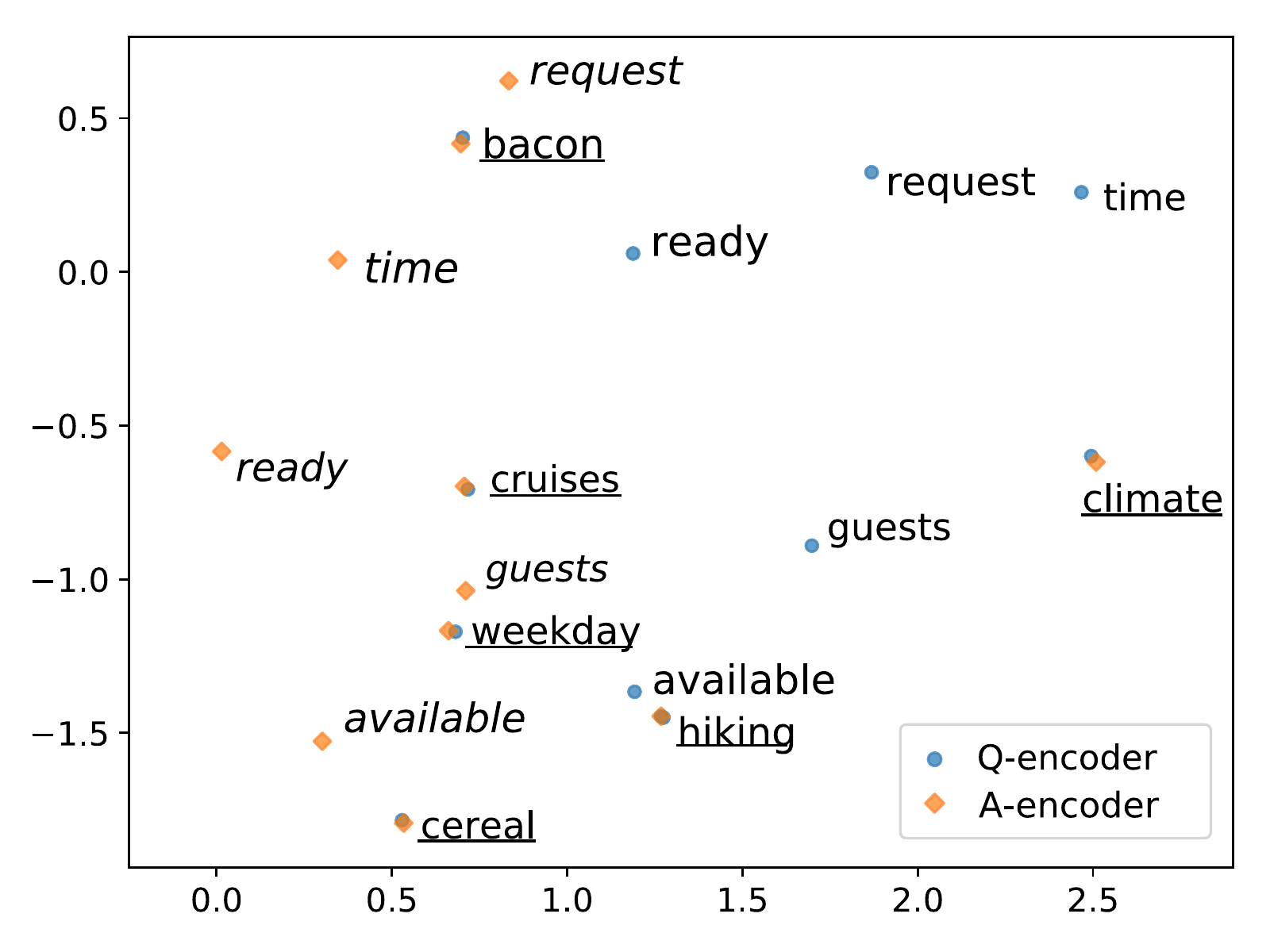}
\caption{Example embedding vectors from question and answer encoders. Underlines denote similar embedding vectors in both the encoders.}
\label{fig:embedding}
\vspace{-8pt}
\end{figure}

\noindent
%{\bf Component-wise Analysis~}

{\bf Multiple Encoders:}
%
%% Multiple encoders
Our motivation to use different encoders for questions and answers is based on the assumption that they use different vocabulary and semantics. We found that there were 8k (72k) unique words in questions, 18k (114k) unique words in answers, and the Jaccard coefficient between two vocabulary sets was 0.25 (0.25) in \trustyou{} (\amazonqa{}), indicating that two sources use significantly different vocabulary.
Also, the same word can have different meanings depending on a speaker, and thus such words in the two sources should be embedded differently. To visualize the embedding vectors of common words in \trustyou{}, we mapped them into 2D space using $t$-SNE~\cite{maaten2008visualizing}.  Fig.~\ref{fig:embedding}
shows that
{\it subjective} words that represents speaker’s attitude (e.g., ``ready'', ``guests'', ``time'') had significantly different embeddings in the question and answer encoders. In contrast, {\it objective} words such as menu, or activity (e.g., ``bacon'', ``cruise'', ``weekday'') had similar embeddings although the two encoders do not directly share the embedding parameters. This indicates that multiple encoders not only capture the different meanings in questions and answers but also retain consistent meanings for words that keep the same meanings in the two sources.

% %% Attention
% \yoshi{Remove this paragraph?} We further analyze the alignments of tokens in output sequence and input sequence for a few interesting examples. As shown in \todo{Figure x}, attention does help the decoder focus on tokens in the input that are relevant to the tuple sequence. \nikita{no difference in results. discuss if we should remove these.}
% %

\noindent
{\bf Relevance Scoring:}
% We evaluated the relevance scoring function by comparing with another \neuron{} model that uses HolE~\cite{nickel2016holographic} instead of TransE.
We compared with another \neuron{} model that uses HolE~\cite{nickel2016holographic} for relevance scoring. Both the HolE and TransE models achieved the same precision of 80.7\%, with HolE achieving slightly higher recall (+1.4\%).
%
%\neuron{} with TransE and HolE achieved precision values of 80.7\% and 80.7\% respectively. 
This suggests that incorporating relevance scoring in \neuron{} can robustly improve the extraction accuracy, regardless of the choice of the knowledge embedding method. We also estimated the upper-bound precision by evaluating if the correct tuple was included in the top-500 candidates. The upper-bound precision was 85.0\% on \trustyou{}, indicating that there is still room for improvement on incorporating relevance scoring.

\subsection{Error Analysis}

We examined a random sample of 100 errors shared by all the systems across the tested datasets. Arguably, encoder-decoder models suffer when extracting tuples from long utterances (avg. of 54 tokens), contributing to 43\% of the errors. 34\% of the incorrectly extracted tuples used words that were shared across the two sources. This indicates that the extractor makes errors when resolving ambiguity in tokens.
28\% of the error cases used informal language that is generally difficult for any extractor to understand. We show some examples (1 and 2 in Table~\ref{tab:discussion}) where \neuron{} successfully combined information across two sources and examples (3 and 4 in Table~\ref{tab:discussion}) where it failed.

\renewcommand{\arraystretch}{0.9}% Tighter
\noindent
\begin{table}
\setlength\tabcolsep{1.5pt} % default value: 6pt
\centering
\fontsize{9}{11}\selectfont
\hspace{-2mm}
\begin{tabularx}{0.5\textwidth}{c|l}
\toprule
\multirow{3}{*}{\textbf{1}}
& {\bf Q}: Tell me what the username and password is for WiFi\\
& {\bf A}: Absolutely! Both the username and passcode is C800.\\
& {\bf StanfordIE}: $\langle$ passcode, is, C800 $\rangle$\\
& {\bf \neuron}: $\langle$ password, is, C800 $\rangle$ \tabularnewline
\midrule
\multirow{3}{*}{\textbf{2}}
& {\bf Q}:  Do hotel guys have ice?\\
& {\bf A}:  There is an ice machine on first floor lobby. \\
& {\bf StanfordIE}: $\langle$ hotel, do, ice $\rangle$ \\
& {\bf \neuron}: $\langle$ hotel, have, ice machine $\rangle$ \tabularnewline
\midrule
\multirow{3}{*}{\textbf{3}}
& {\bf Q}: Is there a charge for parking a rental car on the property?\\
& {\bf A}:  Self-parking will be \$15 per night. \\
& {\bf StanfordIE}: None \\
& {\bf \neuron}: $\langle$ parking, will, charge $\rangle$ \tabularnewline
\midrule
\multirow{3}{*}{\textbf{4}}
& {\bf Q}: arrange late check out for tomorrow?\\
& {\bf A}:  I have notated a 12 pm check out. Normal check out time\\
& \ \ \ \ \ is at 11 am. \\
& {\bf StanfordIE}: $\langle$ normal check, is at, 11 am $\rangle$ \\
& {\bf \neuron}: $\langle$ check, is at, 11 $\rangle$ \tabularnewline
\bottomrule
\end{tabularx}
  \caption{Examples of successful cases (\textbf{1} and \textbf{2}) and failed cases (\textbf{3} and \textbf{4}) from test data.}
  \label{tab:discussion}
\end{table}

% From the supplemnetary material
{%\color{blue} 
We further examined three different scenarios: a) errors are shared by both \neuron{} and \neuralopenie{}, b) errors are made exclusively by \neuron{}, c) errors are made exclusively by \neuralopenie{}. 
For each scenario, we examined a random sample of 100 errors.
We categorize the different sources of errors and report the results in Table~\ref{tab:error_analysis_supplementary}. As shown, \neuron{} is superior on longer utterances compared to  \neuralopenie{} (54 tokens vs. 49 tokens). However, ambiguity in tokens in the two sources is a concern for \neuron{} because it has the flexibility to interpret the question and answer differently. Not surprisingly, informal utterances are hard to translate for both the systems.}

\section{Case Study - KB Extension}\label{sec:extension}
\begin{figure}
  \center
  \includegraphics[width=1.0\linewidth]{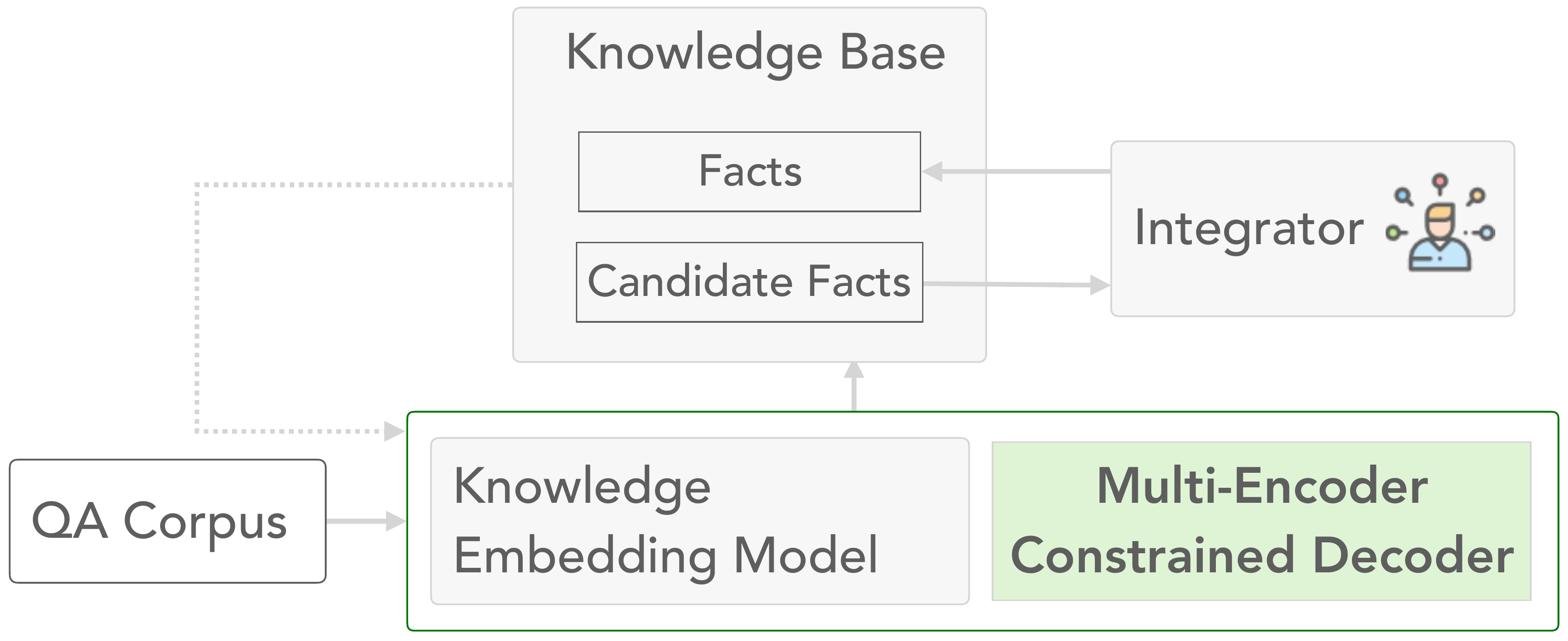}
\caption{Human-in-the-loop system for extending a domain-specific KB.
% \caption{Iteratively extending a KB using tuples extracted by our multi-encoder constrained-decoder and annotated by crowd workers.
% \yoshi{Perhaps, algorithmic procedure with pseudo-code may be easier than figure since each iteration involves not only KB update but also model update.}
}
  \label{fig:kb_extension}
  %\vspace{-0.5em}
\end{figure}

The extracted tuples from \neuron{} can be used to extend a KB for a specific domain. However, automatically fusing the tuples with existing facts in the KB can have limited accuracy. This can be due to noise in the source conversation, no prior knowledge of join rules and more. 
One possible solution is to design a human-in-the-loop system that iteratively extracts tuples and filters them based on human feedback (Fig.~\ref{fig:kb_extension}). In each iteration, a set of tuples is annotated by human annotators based on their relevance to the domain of the KB. 
The tuples marked relevant are added to the KB and the relevance scoring function is updated for extracting more relevant tuples from the corpus in the next iteration.

% Even though effective for KB completion, KB embedding models are purely data-driven. In other words, these models make inferences solely based on existing facts ignoring any logical or physical rules of the output domain~\cite{wang2017knowledge}. 

\renewcommand{\arraystretch}{1.1}
\begin{table}
\centering
\setlength\tabcolsep{1.5pt}
\fontsize{9}{11}\selectfont
	\begin{tabularx}{0.5\textwidth}{|X|l|l|l|}
	    
    	\toprule
    	\rule{0pt}{1.3em}
        	Error Category	&	$\overline{N}$, $\overline{B}$ &	$N$, $\overline{B}$ &$\overline{N}$, $B$ \\
        \toprule
  			long utterances & 43\% & 45\% & 40\%\\
  			\hline
  			avg. length of utterance &	54 tokens	&	49 tokens	&	54 tokens\\
        \midrule
  			ambiguity &	34\%	&	36\%	&	48\%\\
        \midrule
  			informal language &	28\%	&	36\%	&	34\%\\
  		\bottomrule
	\end{tabularx}
  	\caption{Different errors $\overline{N}$ and $\overline{B}$ made by \neuron{} ($N$) and \neuralopenie{} ($B$) respectively.}
  	\label{tab:error_analysis_supplementary}
\end{table}

We conducted a crowdsourced experiment\footnote{\url{https://www.figure-eight.com/}}, simulating the first iteration of the procedure i.e., when no KE model is available. 
We 
%used the FigureEight platform~\cite{FigureEight}
%\footnote{\url{https://www.figure-eight.com/}}
%and 
collected annotations on top-$5$ tuples extracted by \neuron{} for 200 QA pairs in the \trustyou{} dataset. For reliability, we hired five workers for each extraction. The workers were asked to judge if a tuple is relevant to the hotel domain and represents concrete information to be added to a KB. We found precision@5 was 41.4\%, and \neuron{} extracted at least one useful tuple for 83.0\% of the 200 QA pairs. Overall, the system added 243 unique tuples (out of 414 tuples extracted by \neuron{}) to the KB. We also collected annotations for the tuples extracted by \neuralopenie{}. \nikita{The precision@5 and recall@5 values were 41.3\% and 79.0\% respectively. Although the precision values are quite similar, \neuron{} can extract correct tuples from more QA pairs than \neuralopenie{}.}
%
% {\color{blue} We also collected annotations for the tuples extracted by \neuralopenie{}. The precision@5 and recall@5 values were 41.3\% and 79.0\% respectively. Although the precision values are quite similar, \neuron{} can extract correct tuples from more QA pairs than \neuralopenie{}.}
%
While the precision can further be improved, the preliminary results support that \neuron{} is a good candidate for extraction in a human-in-the-loop system for KB extension. We did not use any sophisticated methods for ranking tuples in our experiment. Thus, a better ranking algorithm might lead to improved precision. 
%We leave this for our future work.

%For each iteration, we report the fraction of tuples marked correct (precision) by the workers and the relative size of the KB at the end of the iteration.
% \setlength{\tabcolsep}{2pt}
% \begin{table}
% 	\begin{tabularx}{0.48\textwidth}{|X|l|l|l|l|l}
%     	\toprule
%     	\multirow{2}{*}{Iteration} & \multicolumn{2}{c|}{\trustyou{}} & \multicolumn{2}{c|}{\amazonqa{}}\\
%     	\cline {2-5}
%         & Precision & KB size & Precision & KB size\\ 
%         \toprule
%         	0	& ?	&	1 & ?	&	1\\
%             \hline
%             1	& ?	&	? & ?	&	?\\
%             \hline
%             2	& ?	&	? & ?	&	?\\
%             \hline
%             3	& ?	&	? & ?	&	?\\
%             \hline
%             4	& ?	&	? & ?	&	?\\
%             \hline
%             5	& ?	&	? & ?	&	?\\
%   		\bottomrule
% 	\end{tabularx}
%   	\caption{Precision of top-k tuples from \amazonqa{} and the relative size of KB at each iteration.}
%   	\label{tab:crowdsourcing}
% \end{table}

\section{Related Work}\label{sec:relatedwork}
%The first \openie{} system \textsc{\small TextRunner} ~\cite{yates2007textrunner} modeled extraction as a sequence labeling task. %Since then several \openie{} methods~\cite{fader2011identifying,schmitz2012open,del2013clausie,bhutani2016nested}, using hand-crafted or learned patterns, have attempted to improve the diversity of patterns, the scalability of extraction and support for specific entities/relations. More recently, an end-to-end \openie{} approach~\cite{cui2018neural} was proposed that used an encoder-decoder framework to generate high-quality tuples from sentences without any hand-crafted linguistic/syntactic patterns. \cite{stanovsky2018supervised} uses the sequential tagging approach to train a model for \openie{}. \todo{(TODO: add 1-2 sentence to clarify the difference.)}

There is a long history of \openie{} systems for extracting tuples from plain text. They are built on hand-crafted patterns over an intermediate representation of a sentence (e.g., POS tags ~\cite{yates2007textrunner,fader2011identifying}, dependency trees~\cite{bhutani2016nested,schmitz2012open}). Such rule-based systems require extensive engineering when the patterns become diverse and sparse. Recently, \openie{} systems based on end-to-end frameworks, such as sequence tagging~\cite{stanovsky2018supervised} or sequence-to-sequence generation~\cite{cui2018neural}, have been shown to alleviate such engineering efforts. However, all these systems focus on sentence-level extraction. We are the first to address the problem of extracting tuples from question-answer pairs.

Our proposed system is based on an encoder-decoder architecture, which was first introduced by~\citeauthor{cho2014learning}\ for machine translation. Attention mechanisms ~\cite{bahdanau2014neural,luong2015effective} have been shown to be effective
for mitigating the problem of poor translation performance on long sequences. Their model can learn how much information to retrieve from specific parts of the input sequence at decoding time. There is abundant research on generalizing such frameworks for multiple tasks, specially by employing multiple encoders. Using multiple encoders has been shown to be useful in mutli-task learning~\cite{luong2015multi}, multi-source translation~\cite{zoph2016multi} and reading comprehension~\cite{xiong2016dynamic}. We are the first to explore a multi-source encoder-decoder architecture for extracting tuples from \cqa{} datasets. 

Traditional encoder-decoder architectures are not tailored for information extraction and knowledge harvesting. To make them suitable for information extraction, the sequence generation must be subjected to several constraints on the vocabulary, grammar etc. Recently, grammar structures have been integrated into encoder-decoder models~\cite{iyer2017learning,zhang2017constrained}. There are variations such as Pointer Networks~\cite{vinyals2015pointer} that yield a succession of pointers to tokens in the input sequence. All these studies share a common idea with our paper, which is to enforce constraints at sequence generation time. 
Since we focus on extraction from \cqa{} datasets, our work is broadly related to the literature on relation extraction~\cite{savenkov2015relation,hixon2015learning,Wu:2018:WSDM} and ontology extraction~\cite{kumar2018enriching} from community generated question-answer datasets. However, we differ in our underlying assumption that the relations and entities of interest are not known in advance. \nikita{Alternatively, a \cqa{} dataset could be transformed into declarative sentences \cite{Demszky:2018:transforming} for a conventional \openie{} system. However, such a two-stage approach is susceptible to error propagation. We adopt an end-to-end solution that is applicable to generic \cqa{} datasets.}

% To be discussed (too long?)

% {\color{blue} (Yoshi: too long, will elaborate it) Recently, \cite{Demszky:2018:transforming} has developed a technique that converts QA pairs into the declarative forms. The two-stage approach that consists of the conversion step and conventional \openie{} techniques could be one solution. As shown in the results, we consider an end-to-end framework that directly takes into account question and answers would perform better than the two-stage approach. Also, the scope of \neuron{} is not limited to QA pairs that can be represented as declarative forms, rather can be applied to more general \cqa{} datasets.}

\section{Conclusions and Future Work}\label{sec:conclusion}

We have presented \neuron, a system for extracting structured data from QA pairs for the purpose of enriching knowledge bases. \neuron{} uses a multi-encoder, constrained-decoder framework to generate quality tuples from QA pairs.

\neuron{} achieves the highest precision and recall in extracting tuples from QA pairs compared with state-of-the-art sentence-based models, with a relative improvement as high as 13.3\%.
%by as much as 5.4\%. 
It can discover 15-25\% more tuples which makes it suitable as a tuple extraction tool for KB extension.

There are several directions for future research.  
One interesting direction is to investigate whether \neuron{} can be extended to work on {\em open-domain QA corpus}, which may not be restricted to any specific domain. 
%, we will further investigate frameworks that could take advantage of the input knowledge base to improve the tuple extraction from question-answer pairs. We also plan to explore alternative scoring methods to identify tuples relevant to the domain of the knowledge base.

\section*{Acknowledgement}
\nikita{We thank Tom Mitchell and the anonymous reviewers for their constructive feedback. This work was supported in part by the UM Office of Research.}

\bibliography{naaclhlt2018}
\bibliographystyle{acl_natbib}

% Camera ready can't have Appendix
\clearpage
\appendix

\section{Supplementary Material}
% {\color{red} Following part will be detached as an independent supplemental material which can be submitted accompanied with the main paper though it is optional. We will decide if we submit the supplemental material later.}
This supplementary material contains details of the analysis settings described in Section 4 and additional results not reported in the main paper.

\subsection{Word Embedding Analysis}
We investigated the embedding layers of the question encoder and the answer encoder of the \neuron{} model trained on the \trustyou{} dataset. 

For robust analysis, we ignored non-English words and any words that contained numerical digits (e.g., \#18D, \$10). We used {\textsc{\small pyenchant}}\footnote{v2.0.0 \url{https://github.com/rfk/pyenchant}} for filtering English words. For the remaining words, we find their embedding vectors from the two encoders, concatenate them to create a single matrix. This ensures that same embedding vectors are mapped to the same point in the visualization space. We used $t$-SNE\footnote{ {\textsc{\small TSNE}} v0.20.0 \url{https://scikit-learn.org/stable/} with default configuration} to map embedding vectors into 2D space for visualization.

\iffalse
\subsection{Error Analysis}

We examined three different scenarios: a) errors are shared by both \neuron{} and \neuralopenie{}, b) errors are made exclusively by \neuron{}, c) errors are made exclusively by \neuralopenie{}. 
For each scenario, we examined a random sample of 100 errors.
We categorize the different sources of errors and report the results in Table~\ref{tab:error_analysis_supplementary}. As shown, \neuron{} can translate longer sequences correctly better than \neuralopenie{} (54 tokens vs. 49 tokens). However, ambiguity in tokens in the two sources in a concern for \neuron{} because it has the flexibility to interpret the question and answer differently. Not surprisingly, informal utterances are hard to translate for both the systems.
\fi

\subsection{Crowdsourced Evaluation}
\begin{figure*}[ht]
  \center
  \includegraphics[width=0.9\linewidth]{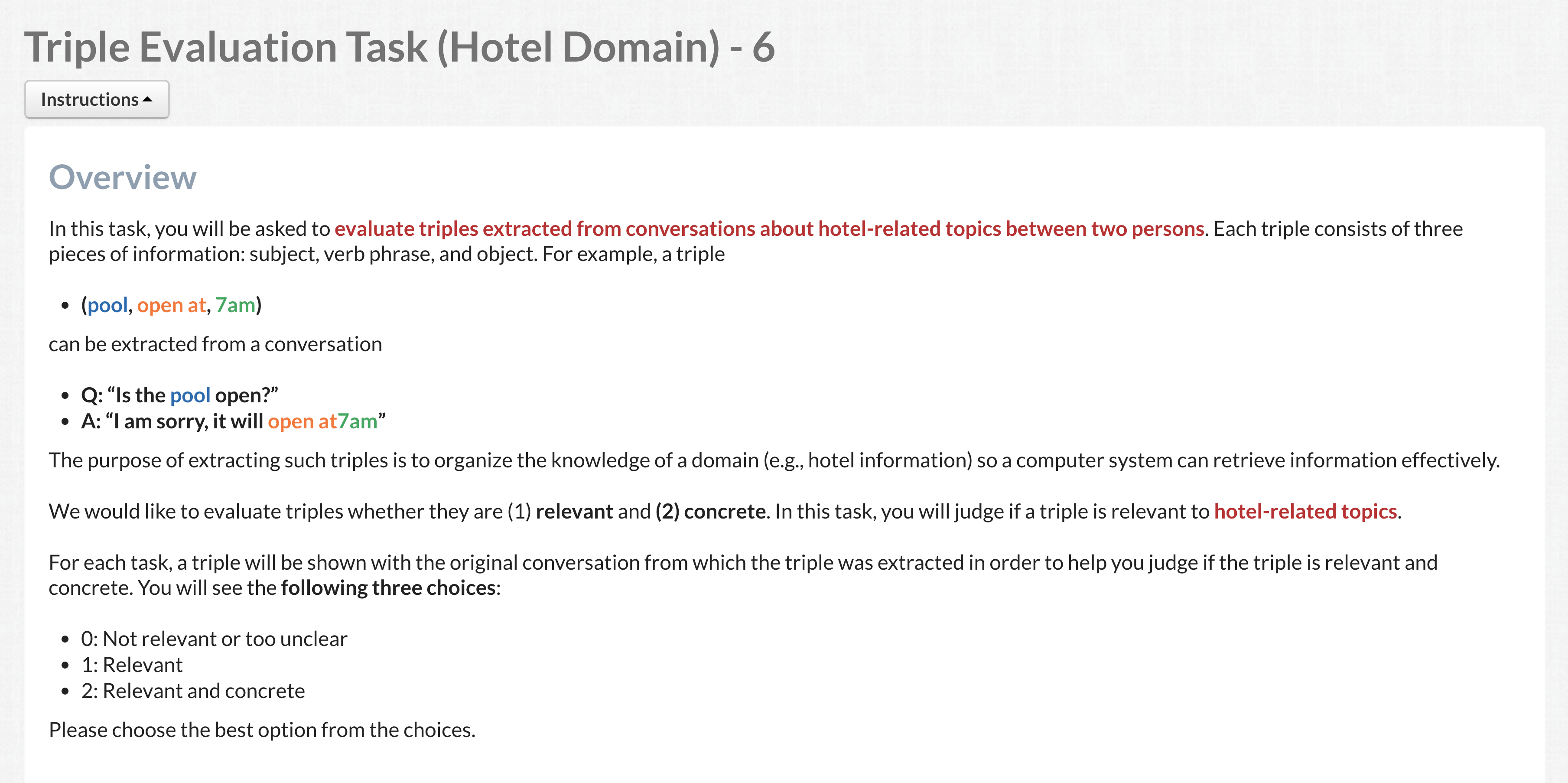}
  \includegraphics[width=0.9\linewidth]{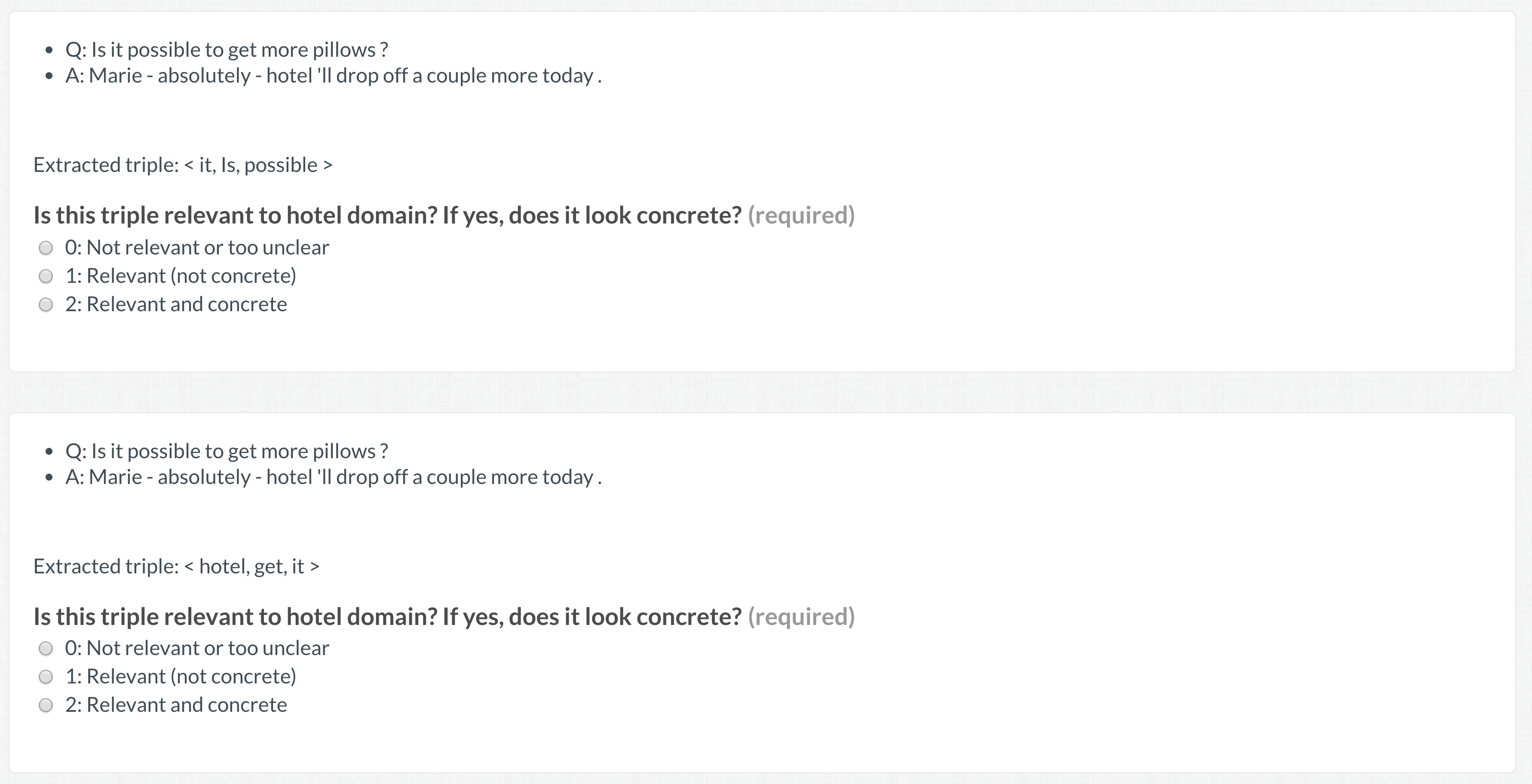}
  \caption{Screenshot of the instructions and examples of the crowdsourced task.}
  \label{fig:crowdtask}
\end{figure*}

Figure \ref{fig:crowdtask} shows the instructions and examples of the crowdsouced task. In the crowdsourcing task, crowdsourced workers were asked to judge after reading an extracted tuple with the original QA pair. Since it is difficult to define the usefulness of the tuples without assuming a KB, we used {\it relevance} and {\it concreteness} as criteria to grade extracted tuples. Specifically, each worker was asked to choose one option from the three options: {\tt Not relevant or unclear} (0), {\tt Relevant} (1), {\tt Relevant and concrete} (2). 

We set \$0.05 as payment for each annotation. We carefully created 51 test questions which were used to filter out untrusted judgments and workers. The platform increases the number of annotators so each tuple should always have 5 trusted annotators. The 5 annotations for each tuple were aggregated into a single label with a confidence value that takes into account the accuracy rates of the annotators based on the test questions.

\iffalse
\renewcommand{\arraystretch}{1.1}
\begin{table}
\centering
\setlength\tabcolsep{1.5pt}
\fontsize{9}{11}\selectfont
	\begin{tabularx}{0.46\textwidth}{|X|l|l|l|}
	    
    	\toprule
    	\rule{0pt}{1.3em}
        	Error Category	&	$\overline{N}$, $\overline{B}$ &	$N$, $\overline{B}$ &$\overline{N}$, $B$ \\
        \toprule
  			long utterances & 43\% & 45\% & 40\%\\
  			\hline
  			avg. length of utterance &	54 tokens	&	49 tokens	&	54 tokens\\
  		\midrule
        \midrule
  			ambiguity &	34\%	&	36\%	&	48\%\\
  		\midrule
        \midrule
  			informal language &	28\%	&	36\%	&	34\%\\
  		\bottomrule
	\end{tabularx}
  	\caption{Different errors $\overline{N}$ and $\overline{B}$ made by \neuron{} ($N$) and \neuralopenie{} ($B$) respectively.}
  	\label{tab:error_analysis_supplementary}
\end{table}
\fi

%%
% - Design
% - Guideline
% - Payment
%%

\end{document}